\documentclass[letterpaper]{article} 
\usepackage{aaai23}  
\usepackage{times}  
\usepackage{helvet}  
\usepackage{courier}  
\usepackage[hyphens]{url}  
\usepackage{graphicx} 
\usepackage{natbib}  
\usepackage{caption} 
\usepackage[algoruled,vlined,linesnumbered]{algorithm2e}

\title{Reducing Redundant Work in Jump Point Search}
\author {
Shizhe Zhao,
Daniel Harabor,
Peter J. Stuckey
}
\affiliations {
  Monash University\\
  \{shizhe.zhao,daniel.harabor,peter.stuckey\}@monash.edu
}

\usepackage{array}
\usepackage{subcaption}
\usepackage{comment}
\usepackage{amsmath, amssymb,amsthm}
\usepackage[textsize=scriptsize,textwidth=1cm]{todonotes}
\usepackage{mathtools}
\usepackage{adjustbox}
\usepackage{cleveref}
\usepackage{booktabs}
\usepackage{placeins}

\usepackage{tikz}
\usetikzlibrary{patterns, matrix, fit, patterns, arrows, calc, positioning, backgrounds}

\theoremstyle{definition}
\newtheorem{definition}{Definition}

\theoremstyle{remark}

\tikzset{coord matrix/.style={
    matrix of nodes,
    column sep=-\pgflinewidth, row sep=-\pgflinewidth,
    nodes in empty cells,
    row 1/.append style={nodes={draw=none}},
    column 1/.append style={nodes={draw=none}},
    nodes={draw,
      minimum height=#1,
      anchor=center,
      text width=#1,
      align=center,
      inner sep=0pt
    },
  },
  coord matrix/.default=0.8cm
}

\tikzset{square matrix/.style={
    matrix of nodes,
    column sep=-\pgflinewidth, row sep=-\pgflinewidth,
    nodes in empty cells,
    nodes={draw,
      minimum height=#1,
      anchor=center,
      text width=#1,
      align=center,
      inner sep=0pt
    },
  },
  square matrix/.default=0.8cm
}

\newcounter{ex}
\newenvironment{example}{\refstepcounter{ex}\begin{trivlist}\item\textbf{Example \theex}}{\end{trivlist}}

\newcommand{\ignore}[1]{}

\begin{document}

\maketitle

\begin{abstract}
JPS (Jump Point Search) is a state-of-the-art optimal algorithm for 
online grid-based pathfinding. 
Widely used in games and other navigation scenarios, JPS nevertheless
can exhibit pathological behaviours which are not well studied: 
(i) it may repeatedly scan the same area of the map to find successors;
(ii) it may generate and expand suboptimal search nodes.
In this work, we examine the source of these pathological behaviours, show 
how they can occur in practice, and propose a purely online approach, called \emph{Constrained JPS} (CJPS), to tackle them efficiently.
Experimental results show that CJPS has low overheads and is often faster than JPS 
in dynamically changing grid environments:  by up to 7x in large game maps 
and up to 14x in pathological scenarios.
\end{abstract}

\section{Introduction}
Grid Based Pathfinding is a classic problem in AI and widely used in games and robotics, as well as
an active research area. 
Despite a variety of fast preprocessing-based approaches for solving this problem~\cite{dblp:conf/socs/sturtevantttuks15},
online approaches are still preferable in dynamic environments where \emph{obstacles may appear (or disappear) on the map between different queries}. 
The reason is that preprocessing-based approaches rely on precomputed auxiliary data structures, which have to be rebuilt or repaired when the map changes.
When dynamic changes are frequent and/or affect large regions 
of the map, the online costs of rebuild or repair operations grows quickly 
and the amortized performance of preprocessing-based algorithms becomes worse than 
purely online approaches~\cite{DBLP:conf/socs/MaheoZAHS021,hechenberger2020online}.

Jump Point Search (JPS)~\cite{harabor2011online,dblp:conf/aips/haraborg14} is a
state-of-the-art algorithm for online pathfinding on uniform-cost grids.
It applies a local pruning policy to prune the successor generation of A$^*$ to reduce symmetries.
Experiments show that JPS is two orders of magnitude faster than A$^*$. 
However, it may suffer from some pathological behaviours. These behaviours cause it to be inefficient in certain topologies of maps that are common in dynamic environments. 
In further sections, we will provide technical details of JPS, and discuss these pathological behaviours.

In this work we examine the root cause of JPS's pathological behaviours.
We then consider how to resolve these issues on-the-fly using a new reasoning technique based on geometric constraints. 
This approach reduces redundant work and helps to avoid the generation and expansion of suboptimal successors. 
We then integrate these constraints with block-based scanning to produce a 
new state-of-the-art algorithm called {\em Constrained JPS} (CJPS).
Finally we conduct an experimental comparison on benchmarks drawn from real
applications and on synthetic setups intended to produce pathological
behaviour. Results show that when the environment is dynamic CJPS can be up to 7x
faster than JPS in terms of cumulative runtime. In pathological setups, 
this improvement can be up to 14x. When there is little or no redundant work,
CJPS has a small overhead of $\approx$ 25\%. Overall, \emph{Constrained JPS} 
represents a substantial advancement for optimal and online grid-based pathfinding 
in dynamic environments.

\section{Background}

In this section we give a brief description of the dynamic grid-based 
pathfinding problem. We then review related works for dynamic environments, 
to define our research scope. We also give a summary of the JPS algorithm,
to help readers understand the weaknesses of the current approach.

\subsection{Problem Definition}
The gridmaps we consider  in this paper are 8-connected, with move directions:
\emph{N,S,W,E,NW,NE,SW,SE}.
Each cell in the map is either traversable or blocked. 
Our vertices $\mathbf{V}$ are 
located at the coordinates of the traversable cells.
Edges $\mathbf{E}$ are defined from vertex to adjacent vertex using one of the move directions.
\textbf{Corner-cut} is not allowed, i.e. in a valid diagonal move, adjacent cells in both of the component cardinal move directions 
(which together comprise the diagonal vector), must also be traversable; 
e.g., in Figure~\ref{fig:jps-pruning-b},
the agent cannot move from $4$ to $2$ as the north cell $1$ is blocked.
For $e\in\mathbf{E}$, let $\mathbf{w(e)}$ be the cost of each move,
all \textbf{cardinal moves} (\emph{N,S,W,E}) have cost $1$, and all \textbf{diagonal moves} (\emph{NW,NE,SW,SE}) have cost $\sqrt{2}$.
A \textbf{path} is represented by $p=[ v_0, v_1, \ldots, v_k ]$, where $v_i \in \mathbf{V}$ and $(v_i, v_{i+1}) \in \mathbf{E}, 0 \leq i<k$.
The length of a path $p$ is $len(p) = \sum_{i=0}^{k-1} w(v_i, v_{i+1})$. 
The shortest path from $s$ to $t$ is a path with the minimum length.
The octile distance between vertices $a,b$ is denoted by $|ab|$.
We assume the environment can change between pathfinding queries. 
We further assume that these changes are uniformly distributed on the map.

\subsection{Related Works for Dynamic Environment} 
Pre-computed estimators, such as Landmark
Heuristics~\cite{goldberg2005computing}, Differential Heuristics
(DH)~\cite{DBLP:conf/ijcai/SturtevantFBSB09} and CPD
Heuristics~\cite{ijcai2019-167},
have the potential to handle some types of dynamic environments. 
In particular, these approaches remain admissible under the assumption that
dynamic changes online will never decrease the cost of any optimal path 
to less than its cost during the offline phase.
Pre-computed heuristics are orthogonal to JPS, which does not 
come with any cost-based assumptions.
Another approach, Customizable Contraction Hierarchies (CCH)~\cite{dgpw-crp-17} 
is an abstraction-based preprocessing technique which can repair 
its auxiliary data online, after a dynamic change. CCH is fast, optimal and 
it does not rely on any cost-based assumptions. 
A main disadvantage is that as the number and frequency of map changes increases 
the amortized cost of repair grows large, to the point where it becomes faster to
use a reference algorithm such as A*~\cite{DBLP:conf/socs/MaheoZAHS021}.

\subsection{Jump Point Search}
Symmetry breaking is a main challenge in 2D path finding. This problem occurs when there exist multiple 
shortest paths from start to target, each of which is derived from any of the others by simply 
re-ordering grid moves. JPS eliminates symmetries by exploring only {\em canonical paths}, where 
diagonal moves appear as early as possible.
The search framework of JPS is the same as A$^*$ and uses the same heuristics for prioritising
nodes in the \textsc{Open} list. In this work we use \textbf{octile distance}, a popular
heuristic similar to the Manhattan estimator but which allows diagonal moves.
The main difference between JPS and A$^*$ is the successor function.
In JPS, \emph{local} suboptimal and non-canonical grid neighbours are pruned on the fly using
a series of simple rules which are recursively applied. 
We define these rules and basic concepts of JPS below.

\begin{figure}[bt]
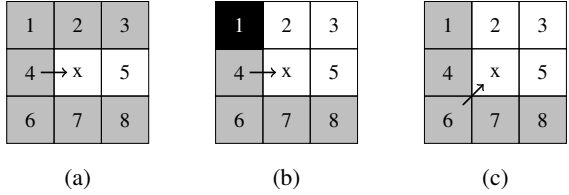

  \begin{subfigure}{.33\linewidth}
    \centering
    \begin{adjustbox}{max width=.75\linewidth}
    \begin{tikzpicture}
      \input{src/jps-pruning.tex}
      \jpsCaseA
    \end{tikzpicture}
    \end{adjustbox}
    \caption{}
    \label{fig:jps-pruning-a}
  \end{subfigure}%
  \begin{subfigure}{.33\linewidth}
    \centering
    \begin{adjustbox}{max width=.75\linewidth}
    \begin{tikzpicture}
      \input{src/jps-pruning.tex}
      \jpsCaseC
    \end{tikzpicture}
    \end{adjustbox}
    \caption{}
    \label{fig:jps-pruning-b}
  \end{subfigure}%
  \begin{subfigure}{.33\linewidth}
    \centering
    \begin{adjustbox}{max width=.75\linewidth}
    \begin{tikzpicture}
      \input{src/jps-pruning.tex}
      \jpsCaseB
    \end{tikzpicture}
    \end{adjustbox}
    \caption{}
    \label{fig:jps-pruning-c}
  \end{subfigure}
  \caption{
    JPS pruning example. In all figures: $x$ is the current node, arrow shows the incoming direction,
    black cells are blocked, white cells are successors and gray cells are pruned.
  }
  \label{fig:jps-pruning}
\end{figure}

\begin{definition}\label{def:diag-first}
  Let $NB(x) = \{ v ~|~ (x,v) \in \mathbf{E}\} $ be the adjacent vertices of a current node $x$, and let $p$ be the parent of $x$ during search.
  Define $LP_p^x(t)$ as the set of optimal paths starting at $p$, ending at $t$, and restricted to only use vertices from $NB(x)$.
  We say that $t$ belongs to the successor set of $x$ (coming from $p$) if $\forall l \in LP_p^x(t)$:
  \begin{equation}
    len([p, x, t]) = len(l) \land rank([p, x, t]) \leq rank(l) 
  \end{equation}
   \noindent 
  Where $rank(\pi)$ is a function that returns the index of the first diagonal move of a path $\pi$.
\end{definition}

\begin{example}
  In Figure~\ref{fig:jps-pruning-a}, let $x$ be the current search node that comes from node $4$. $LP_{4}^{x}(2)=\{[4,2]\}$, and $[4,x,2] \notin LP_4^x(2)$, so 2 is not a successor.
  $LP_4^x(3)={[4,2,3], [4,x,3]}$, $[4,x,2] \in LP_4^x(3)$ but the diagonal move in $[4,2,3]$ appears earlier than the diagonal move in $[4,x,3]$,
  i.e., $rank([4,x,3]) > rank([4,2,3])$, so 3 is not a successor.
\end{example}

\paragraph{Concepts:} we refer to the successor set of node $x$ with the notation $neib(\Vec{d}, x)$, where $\Vec{d}$ is the incoming direction inferred from $p$ and $x$.
These are sometimes called the \textbf{diagonal-first} (equiv. canonical) neighbours of node $x$ and they can be computed on-the-fly in constant time. 
When moving in a straight direction $\Vec{d}$, JPS often reduces the pruned successor set to size 1, i.e., $|neib(\Vec{d}, x)|=1$ (see Fig~\ref{fig:jps-pruning-a}). 
Rather than adding such nodes to the \textsc{Open} list JPS immediately expands them, thus applying the pruning rules anew in a recursive fashion. 
The recursion stops when reaching a node $x'$ where $|neib(\Vec{d}, x')| = 0$ (\textbf{dead-end}) 
or when $|neib(\Vec{d}, x')| > 1$ (\textbf{jump point}), which can occur due to adjacent obstacles (see Fig~\ref{fig:jps-pruning-b}). 
Recursing across the grid looking for jump point successors is called \textbf{scanning}. 
In the second case $x'$ is generated as a {\bf jump point} successor of $x$.
A vertex $v$ is a \textbf{corner point} if $\exists d \in \{N,E,W,S\}: |neib(\Vec{d}, v)|>1$.
When moving in a diagonal direction $\Vec{d'}$, JPS first scans in each of the two corresponding cardinal directions, 
looking for jump points. These are immediately added to the $\textsc{Open}$ list 
and then JPS takes one diagonal step further on the grid (see Fig\ref{fig:jps-pruning-c}). 
This procedure is called \textbf{diagonal recursion} and it continues until the next move $\Vec{d'}$ is a dead-end or corner-cut.

\begin{figure}[bt]
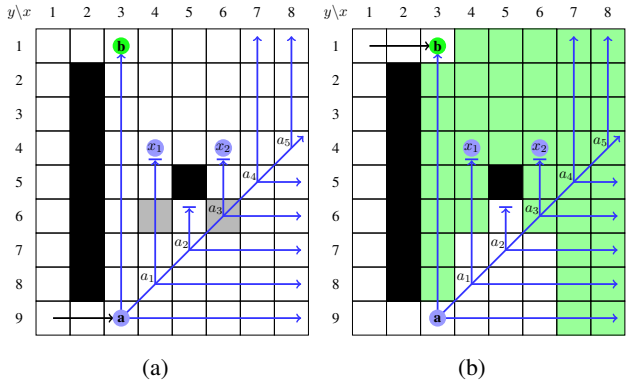

  \begin{subfigure}{.5\linewidth}
    \begin{adjustbox}{max width=\linewidth}
      \begin{tikzpicture}
        \input{src/redundant-work.tex}
        \scanA
      \end{tikzpicture}
    \end{adjustbox}
    \caption{}
    \label{fig:jps-scanning}
  \end{subfigure}%
  \begin{subfigure}{.5\linewidth}
    \centering
    \begin{adjustbox}{max width=\linewidth}
      \begin{tikzpicture}
        \input{src/redundant-work.tex}
        \scanB
      \end{tikzpicture}
    \end{adjustbox}
    \caption{}
    \label{fig:redundant-work}
  \end{subfigure}
  \caption{
    In (\subref{fig:jps-scanning}),
    node $a$ is the current search node, the black arrow represents the coming direction,
    blue arrows represent scanning, $a_{i \in 1\ldots 5}$ are start points of each cardinal scanning in diagonal, 
    gray nodes are corner points, $b, x_1,x_2$ are identified successors (jump points).
    In (\subref{fig:redundant-work}), $b$ is another search node that performs a symmetric diagonal recursion, and green nodes are scanned during this procedure.
  }
\end{figure}

\begin{example}\label{exp:jps-scan}
  In Figure~\ref{fig:jps-scanning}, node $a$ is a jump point with successors (diagonal first neighbour set) $\{$N, E, NE$\}$.
  JPS first scans in directions N and E, and finds a jump point $b$ to the north. 
  Then it moves one diagonal step to $a_1$ and applies the same scanning, finding $x_1$ at the node above $a_1$ during the North scan.
  Similarly, on the North scan, JPS will find a dead-end from $a_2$, and a jump point from $a_3$. 
  Grey cells are corner points that wouldn't stop the scanning.
  JPS keeps moving diagonally and scanning cardinally until the diagonal direction is blocked. 
\end{example}


\paragraph{Block-Based Scanning.}
\emph{Block-based scanning}~\cite{dblp:conf/aips/haraborg14} represents the traversability of the gridmap by a bitmap,
i.e. 0/1 means traversable/non-traversable; then instead of checking node by node, we can load a block of bits for adjacent columns/rows into memory, 
and quickly compute the position of the first jump point in this direction if there is one, 
by using three bitwise operations.
This procedure is branch-less and leverages SIMD instructions which can be extremely fast,
it improves average times of JPS by nearly one order of magnitude.
The only overhead is we need two bitmaps, one stored rowwise and one stored columnwise. 
For more details see example~\ref{exp:block-scan}.
\begin{example}\label{exp:block-scan}
  Assuming the block size is 3, in Figure~\ref{fig:jps-pruning-b}, the bit string of each row (`1,2,3',`4,x,5' and `6,7,8') are $s_1=100_2$, $s_2=000_2$, $s_3=000_2$.
  In a left-right travel, we can quickly compute the first position that bits change from $1$ to $0$ by bitwise operation on $s_1$, 
  which indicates a traversable node next to an obstacle;
  apply same reasoning on $s_2$ and $s_3$, we will find that $x$ is a jump point.
\end{example}

\section{Pathological Behaviours in JPS}\label{sec:motivation}

Known for its fast performance and strong optimality guarantees, JPS nevertheless 
suffers from two distinct types pathological behaviour, each of which can create
substantial amounts of redundant work. 

{\bf Pathological Behaviour \#1:}
JPS may scan the entire map multiple times while computing successors during search.
In~\cite{dblp:conf/ijcai/sturtevantr16} the authors tackle this issue by adding
a scan limit during the successor generation step of JPS. 
The resulting algorithm, known as Bounded JPS, introduces a successor node
each time the limit is reached. 
This approach mitigates redundant scanning but at the cost of 
additional heap operations on the \textsc{Open} list. 

Moreover practitioners must carefully choose a suitable limit, usually for each 
map and potentially for each query, in order to achieve a performance improvement.
A related algorithm, Boundary Lookup JPS~\cite{dblp:journals/tciaig/traishtm16},
replaces the linear scanning operation of JPS with a binary search.
This approach has better complexity but is much slower in practice than
optimised JPS implementations, which rely on an instruction-level parallelism
technique known as block-based scanning~\cite{dblp:conf/aips/haraborg14}.

{\bf Pathological Behaviour \#2:}
JPS may generate and expand suboptimal search nodes. 
This behaviour has not been reported previously in the literature
but could be a cause for concern --- if many suboptimal search nodes are created and 
expanded, there is a potentially large overhead for JPS vs. A$^*$, 
since A$^*$ only expands each node at most once (assuming with a consistent heuristic). 
It is therefore important to understand how suboptimal expansions can occur in JPS, 
how much performance is affected by this behaviour and how to avoid it
in practice.

\begin{example}\label{ep:redundant_work}
In Figure~\ref{fig:redundant-work}, expanding nodes $a$ and $b$ produces 
overlapping diagonal recursions where nodes are scanned multiple times. 
Depending on the g-value of $a$ and $b$, some of the resulting successor nodes
may be suboptimal. For example, when $g_a = g_b$, $x_1, x_2$ are bettered reached from $b$, but JPS still generates them as successors of $a$.
  Furthermore, suboptimal nodes $x_1, x_2$ will be expanded later.
\end{example}

\section{Constrained JPS}\label{sec:cjps}

\paragraph{Observation} In Example~\ref{ep:redundant_work}, scan from $a$ will stop at $b$ (as it is a jump point to $a$), and the same applies to $b$. 
Therefore, $a$ and $b$ have a chance to know each other's g-value during their diagonal recursion, 
which can be utilized to reduce redundant work. 

Based on the observation above, we propose \emph{Constrained JPS}. 
At a high level, when we expand node $a$ and the cardinal scanning finds 
a jump point $v$ with an existing $g$ value, we can create a constraint for the same cardinal direction during the current diagonal recursion.
This constraint restricts cardinal scanning in the later diagonal recursion by a dynamic jump limit, 
and it can be deleted or updated during the recursion. 
We first give a formal definition of the constraint and then we demonstrate how to compute it, when to delete or update it and how to use it for pruning.

\begin{definition}\label{def:constraint}
  A \emph{constraint} is defined by a tuple $\langle a, v, \Vec{d}, L \rangle$, where:
  \begin{itemize}
    \item $a$ is a vertex which starts or continues a diagonal recursion;
    \item $\Vec{d}$ is the cardinal scanning direction in the diagonal recursion;
    \item $v$ is a jump point found by a scanning in direction $\Vec{d}$ from $a$, which has an associated g-value (i.e., it has already been included in the OPEN list or has been expanded);
    \item $L$ is an integer that indicates the maximum number of recursive diagonal moves from $a$.
      A constraint is \textbf{applicable} if the number of diagonal moves from $a$ to $a_i$ is not greater than $L$.
  \end{itemize}
 For the remainder of this section, we use $i$ for the $i$-th diagonal move from $a$ and $a_i$ for the corresponding vertex.
\end{definition}

\begin{figure}[t]
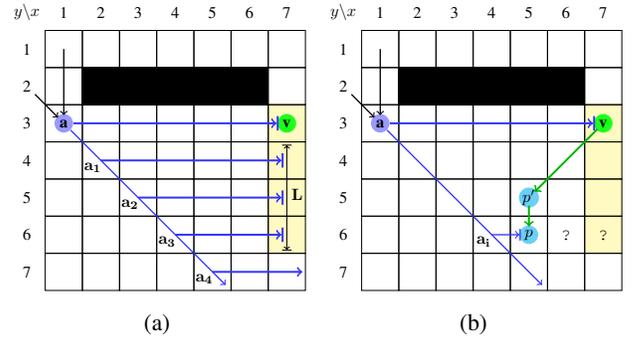

  \begin{subfigure}{.5\linewidth}
    \centering
    \resizebox{\linewidth}{!}{
      \begin{tikzpicture}
        \input{src/cjps-def.tex}
        \constraint
      \end{tikzpicture}
    }
    \caption{}
    \label{fig:cjps-block}
  \end{subfigure}%
  \begin{subfigure}{.5\linewidth}
    \centering
    \resizebox{\linewidth}{!}{
      \begin{tikzpicture}
        \input{src/cjps-def.tex}
        \estimate
      \end{tikzpicture}
    }
    \caption{}
    \label{fig:cjps-estimate}
  \end{subfigure}
  \caption{
    (\subref{fig:cjps-block}) shows constraint $(a, v, d, L)$, where
    $a$ is a jump point coming from north, or a continued diagonal move from northwest; 
    $v$ is a previously visited node that found from the cardinal scanning from $a$, 
    and $L$ is the maximum number of steps where the constraint is applicable,
    yellow cells are better reached from $v$, thus cardinal scanning from $a_1,a_2,a_3$ is stopped by the constraint at the yellow blockage;
    (\subref{fig:cjps-estimate}) shows how to evaluate $\overline{g}_p$, where $p$ is the stop location of a scanning within the constraint.
    The traversability of all nodes after $p$ is unknown, denoted by `\texttt{?}'.}
  \label{fig:cjps-exp}
\end{figure}

$L$ guarantees that when a constraint is applicable, the cardinal scanning in direction $\Vec{d}$ on $a_i$ shouldn't take more than $|av|-i$ steps.
Such a restriction is a perpendicular blockage for scanning in direction $\Vec{d}$ and any path crossing the blockage
is better reached from $v$. Figure~\ref{fig:cjps-block} illustrates the idea. 
We now show how to compute $L$.

\paragraph{Computing $L$}
Let $g_a$ be the g-value of node $a$ and $g_v$ be the g-value of node $v$ found when expanding $a$. 
To compute $L$, we need to consider the following cases:

(i) If $g_a + |av| \leq g_v$, all scans during the diagonal recursion at $a$ will cross the blockage 
with a better g-value than from node $v$; no constraint is applicable, so we set $L=0$.

(ii) If $g_a + \sqrt{2}|av| > g_v + |av|$, 
the diagonal recursion from $a$ should be terminated after $i=|av|$ diagonal steps;
since all further nodes are better reached from $v$ we let $L=|av|$;

(iii) Otherwise $L$ is the minimum integer satisfying:
$$
\left\{ \begin{array}{ll}
    L \geq 0 \land L \leq |av| \; \land\\
    \overbrace{g_a + \sqrt{2}L + (|av|-L)}^\text{distance from a} < \underbrace{g_v + L}_\text{distance from v} \\
  \end{array} \right.
  $$

\paragraph{Updating and Deleting Constraint}
When the constraint is applicable, at the $i$-th diagonal step (i.e., at $a_i$), there are two cases in the cardinal scanning:
i) the scan stops at the blockage; ii) the scan is stopped by a jump point or a dead end 
at node $p$, before reaching the blockage.
In the first case, we can still use the constraint as long as it is applicable.
In the second case, we must discard the current constraint and create a new constraint on $p$. 
To do this, we need to estimate a tight upper bound for the g-value of $p$:
\begin{equation}
  \overline{g}_p = \min\{g_p, |vp'| + 1\} \label{eq:gp}
\end{equation}
where $g_p$ is the $g$ value currently stored with $p$ (or $+\infty$ if it has no stored value),
and $p'$ is the cell in the previously scanned row/column adjacent to $p$. 
This bound is safe since the previous scans of rows/columns from $v$ to $p'$ 
must be empty up to the blockage, since we haven't yet deleted the constraint from $v$.
Therefore, we update the constraint to $\langle a_i, p, d, L'\rangle$, 
where $L'$ is computed (as discussed previously) 
based on $g_{a_i}, \overline{g}_p$ and $|a_ip|$,
Figure~\ref{fig:cjps-estimate} gives an example. 
When $L'=0$, the new constraint is not applicable, then we can simply delete it.

\paragraph{Pruning and Early Termination}
Equation~\ref{eq:gp} can estimate whether node $p$ is better reached from $v$ than $a$.
When $p$ is a jump point, we can prune it if:
$$
\overline{g}_p < g_{a_i} + |a_ip|
$$
and when $p$ is at the blockage, we can terminate the diagonal recursion early if:
$$
\overline{g}_p + |a_ip| < g_{a_i}
$$
Algorithm~\ref{alg:cjps} illustrates the horizontal constraint of CJPS, 
the vertical constraint is imposed similarly. In practice, both can be applied orthogonally in the same diagonal recursion.

\begin{figure}[t]
  \centering
  \resizebox{.62\linewidth}{!}{
    \begin{tikzpicture}
      \input{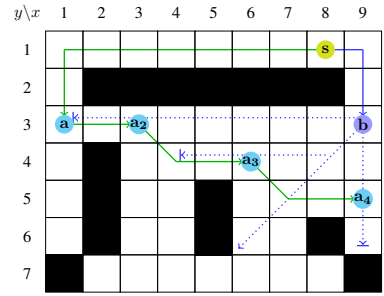}
      \drawcase
    \end{tikzpicture}
  }
  \caption{$s$ (yellow) is the source node, $a$ (cyan) and $b$ (blue) are successors of $s$. 
    $a_2,a_3,a_4$ are suboptimal nodes propagated from $a$,
    their optimal parent is $b$, but the scanning from $b$ won't stop at them.
  }
  \label{fig:subopt_expd2}
\end{figure}

\begin{algorithm}[t]
  \small
  \SetKwInput{Input}{Input}
\SetKwFunction{calcL}{calcL}
\SetKwFunction{calcG}{calcG}
\SetKwFunction{scan}{scan}
\Input{
  $n=(x, y)$: the location of current search node; \\
  $d=(dx, dy)$: the direction of the diagonal move; \\ 
  $d_h=(dx, 0)$: the direction of horizontal scanning; \\
  $c_h=(a, v, d_h, L)$: the constraint for horizontal scanning; \\
  $db$: the bit map for block based scanning; \\

\calcL{$g_a, g_v, |av|$}{
  {:} computing $L$ based on definition~\ref{def:constraint};\\
}
\calcG{$v, p$}{
  {:} computing upper bound of $g_p$ based on equation~\ref{eq:gp}; \\
}
\scan{$n, d, JL, db$}{
  {:} blocked based scanning at $n$ in $d$ with the jump limit $JL$; \\
}
}
$i \gets 0$; \\ 
\While{\textsf{empty}(x+dx, y) and \textsf{empty}(x, y+dy)} {
  $n \gets n+d$; \\
  \If{$i \leq L$} { $i \gets i + 1$;}
  \If{$i \leq L$ and $n$ is better reached from $v$} { break; }
  \If{$i \leq L$} {
    $JL \gets |av|-i$;
  }\Else {
    $JL \gets \infty$;
  }
  $p \gets \textsf{scan}(n, d_h, JL, db)$; \\
  \If{$|np| < JL$} {
    $g_p \gets \textsf{calcG}(v, p)$;\\ 
    $c_h \gets {n, p, d_h, \textsf{calcL}(g_n, g_p, |np|)}$; \\
    $i \gets 0$; \\
  }
  \If{p is jump point and better reached from n} {
    successors.add(p); \\
  }
}

  \caption{
Horizontal constraint in diagonal recursion.
}
  \label{alg:cjps}
\end{algorithm}

\section{Branch-less Implementation}\label{sec:brless}



Block-based scanning can improve JPS by nearly an order of magnitude~\cite{dblp:conf/aips/haraborg14}, thus we need the proposed approach to work with it.
CJPS computes a jump limit on-the-fly based on the information obtained from scanning. 
However, blocked-based scanning is branch-less leveraging SIMD instructions, 
so applying a jump limit is not trivial since adding any if-then-else statement would significantly slow down the scanning.
To force the scanning to stop at certain jump limit in a branch-less way, instead we set an artificial obstacle on the map before the scanning and unset it afterwards.

\section{Eliminating Suboptimal Node Expansion}\label{sec:elimnate_sub}
We have shown that \emph{CJPS} can prune redundant scanning and suboptimal nodes, but it doesn't guarantee to eliminate all of them.
Other online algorithms, e.g. A$^*$ and Dijkstra, are less efficient in practice but guarantee no suboptimal node expansion.
Thus we need to answer the following questions: 
1) how does this happen?
2) how bad could it be?
3) can we eliminate all suboptimal node expansion?

\subsection{How Does It Happen?}
Let $a$ be a non-target search node that is expanded by JPS with suboptimal g-value.
We can infer that node $a$ must be a corner point (cf. jump point) when 
scanning from its optimal parent -- call this optimal direction $\Vec{d}$.
In other words $(\Vec{d}, n)$ does not form not a jump point from the optimal parent. Also, 
since node $n$ is not the target node either, it cannot be generated as a successor when we expanded its optimal parent. 
Notice that the optimal path to the target remains in the JPS search space, but the optimal path to $n$ does not. 
Figure~\ref{fig:subopt_expd2} shows an example. 

We now see that expanding a suboptimal node \emph{v} requires two conditions: 
(i) \emph{v} is reachable from at least two parents (i.e., jump points or the start node);
(ii) \emph{v} does not appear in the search space of its optimal parent due to the JPS pruning rule in Definition~\ref{def:diag-first}. 
This situation is more likely to happen when the map has many corner points and when the heuristic is less accurate; e.g., on game maps.

\subsection{How Bad It Could Be?}

\begin{table}[tb]
  \small
  \centering
\begin{tabular}{lrrrrrr}
  \toprule
  domain    & \#maps & mean & median & sub\% & subp\% \\
  \midrule
  dao       & 154 & 215  & 62   & 30  & 64  \\
  bgmaps    & 75  & 98   & 41   & 27  & 65  \\
  starcraft & 75  & 1038 & 777  & 28  & 73  \\
  street    & 30  & 596  & 312  & 33  & 84  \\
  iron      & 35  & 5065 & 3962 & 43  & 88  \\
  maze512   &  6  & 8863 & 2516 &  0  & --- \\
  random10  &  1  & 7574 & 7527 & 33  & 18 \\
  rooms     &  4  & 1766 &  653 & 25  & 26 \\
  \bottomrule
\end{tabular}

  \caption{
    Statistic of each domain. We count node expansions for the 100 longest queries of each map. 
  \textit{mean, median} are for node expansions, which indicates the difficulty,
    \textit{sub\%} shows the proportion of suboptimal node expansions, 
    and \textit{subp\%} shows the proportion of propagated suboptimal node expansion among \textit{sub}.
}
  \label{tb:subopt_cnt}
\end{table}

The successors of a suboptimal search node may also be expanded, and the
suboptimality may be propagated; e.g. nodes $a_2,a_3,a_4$ in
Figure~\ref{fig:subopt_expd2}. When this happens it is possible that suboptimal
node expansions can dominate the entire search. It is hard to perform a
theoretical analysis on the number of suboptimal node expansions in general
because such behavior depends on many factors, such as the topology of the map,
the order of expansion, and the target location. To understand how frequently
this happens in practice we solve a subset of queries for a variety of 
benchmark domains.
We counted the number of suboptimal node expansions per query and we also
counted the number of propagations; i.e., the number of expanded nodes whose
parent is suboptimal. Table~\ref{tb:subopt_cnt} shows the result. 
We see that, there are no suboptimal nodes in \emph{maze512} domain. This is
due to the underlying tree structure of these maps, which means that search nodes 
are reachable from only one jump point. In other domains,
more than 25\% of node expansions are suboptimal, and many of these are
propagated from their parent. This implies that pruning suboptimal nodes
earlier could avoid expanding more suboptimal nodes in the future. 

\subsection{Can We Eliminate All of Them?}
Based on the discussion above, it is clear that we can eliminate all suboptimal node expansion by storing a $g$-value on every corner point, 
rather than just on every jump point.
However, this approach comes with significant overhead since, for each cardinal scan, we needs to stop at each encountered corner point.
Although this does not change the total number of scanned nodes, it can significantly affect the performance of block-based scanning.
Therefore there is a trade-off, between introducing overhead on the one hand and reducing redundant work on the other.
In this section we explore two ways to mitigate redundant expansions with minimal overhead costs.

\paragraph{Diagonal Caching}
Here we store a g-value on each corner point found during diagonal recursion. 
Since the diagonal recursion stops at every cell on the diagonal, there is little overhead,
mainly arising from an extra memory access during the scan.
However, this approach only allows pruning cells that appear along diagonals from an expanded node, 
and most cells do not meet this condition.

\paragraph{Backwards Scanning}
When a cardinal scan in direction $\Vec{d}$ finds a jump point $n$, 
we start a new scan from $n$ in the reverse direction.
The forward to $n$ can pass at most two corner points before reaching $n$ - 
one for the row above and row below (or the column left and right).
The reverse scan can thus stop after labelling at most 
two corner points (or upon reaching the parent of $n$, whichever comes sooner).
Since not all (forward) scanning finds a jump point (most are dead-end) 
this approach should have smaller overhead than stopping at all corner points.
However it is more expensive than Diagonal Caching.

\section{Experimental Setup}\label{sec:cjps-exp}
We compare \textit{CJPS} and \textit{JPS} on two distinct benchmarks: synthetic maps, 
which shows performance in extreme cases, and domain maps, which shows performance on a range 
of well-established test sets drawn from real applications. Table~\ref{tb:subopt_cnt} shows a summary.
CJPS and JPS are both implemented in C++ and both make use of block-based scanning~\cite{dblp:conf/aips/haraborg14}.
We compile with \emph{clang 13} using \emph{-O3} under \emph{5.10.102-1-MANJARO} and we test on a \emph{Intel Xeon E-2276M} processor 
with \emph{32 GB} RAM. Our implementations and data are available online.\footnote{https://github.com/eggeek/constrained-jps}

{\bf Domain maps:}
From Sturtevant's benchmark set~\cite{sturtevant2012benchmarks} we select all
game domains (bgmaps, starcraft, dao) and all grid rasterisations of real
cities (streets). We also experiment with the Iron Harvest domain, a recent
collection of 35 grid maps taken from the
game~\cite{DBLP:conf/socs/HaraborHJ22}. These maps are much larger than and
much more challenging than other game benchmarks. 

\begin{figure}[t]
  \centering
  \includegraphics[width=\linewidth]{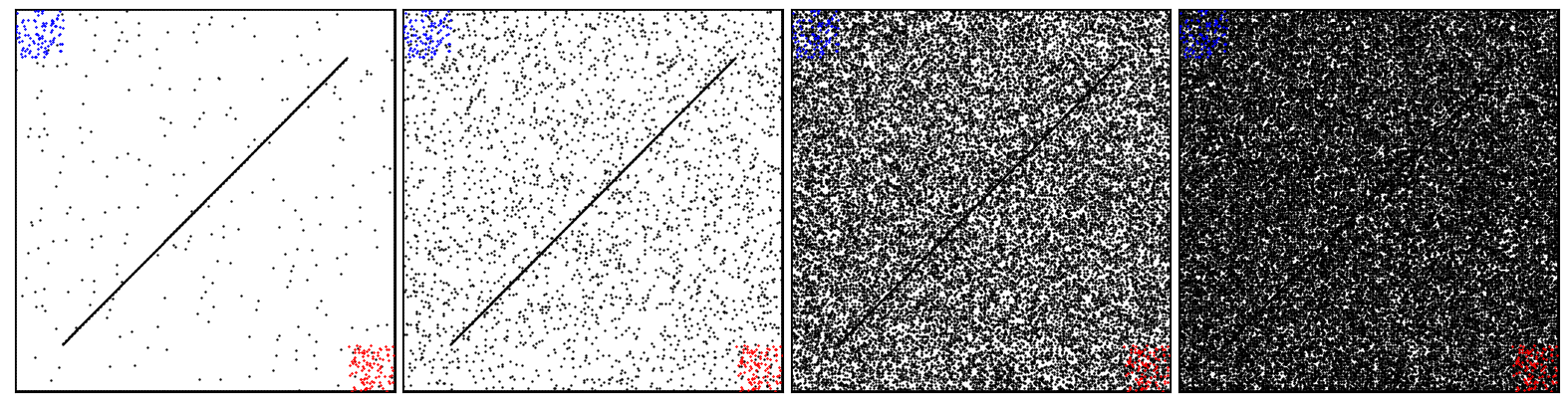}
  \caption{Synthetic maps, where $s=512$, $b=75\%$ and $r \in$ \{0.1, 1, 10, 20\}. Blue and red clusters are starts and targets.}
  \label{fig:visual_map}
\end{figure}

{\bf Synthetic maps:} these are pathological test cases featuring an empty map with random obstacles and a diagonal blockage in the middle.
We control for three variables: 
\begin{itemize}
  \item $r$: the proportion of traversable cells that are blocked, which simulates dynamic environments;
  \item $s$: the height and width of map is $s\times s$;
  \item $b$: the proportion of diagonal blockage, which controls the difficulty, i.e., less accurate heuristic;
\end{itemize}
There are 100 instances per map, where starts and targets are always traversable and clustered in top-left and bottom-right regions.
Figure~\ref{fig:visual_map} shows an example.

\noindent

\section{Results}
For all experiments that measure execution time, we run each map 10 times in random order and choose the median, to avoid cache behaviour and reduce random noise.
For all experiments that measure suboptimal behaviour, we run a Dijkstra to compute a true-distance table before each query. 

\subsection{Exp-1: Synthetic Maps}

\begin{table}[tb]
  \small
  \begin{subtable}{\linewidth}
  \centering
  \begin{tabular}{l|llrr}
  \toprule

   & \multicolumn{4}{c}{Improvement Factors} \\
  r (\%) & hp-opt  & subopt & expd cost         & runtime \\
  \midrule
  0.0    & 1.00 & ---    & 0.79 (5.76/ 7.34) & 0.79       \\
  0.1    & 1.64 & 37.10  & 8.95 (3.24/ 0.36) & 14.87      \\
  1.0    & 1.56 & 3.84   & 4.16 (1.51/ 0.36) & 6.47       \\
  10.0   & 1.15 & 1.44   & 1.03 (0.39/ 0.37) & 1.18       \\
  20.0   & 1.08 & 1.26   & 0.89 (0.30/ 0.34) & 0.96       \\
  \bottomrule
  \end{tabular}
  \caption{Vary obstacle density, fix s=512, b=75\%.}
  \label{tab:vary-r}
\end{subtable}

\begin{subtable}{\linewidth}
  \centering
  \begin{tabular}{l|lllrl}
    \toprule
    & \multicolumn{4}{c}{Improvement Factors} \\
    b (\%) & hp-opt  & subopt& expd cost & runtime \\
    \midrule
    0.00 &   1.00 &    1.00 &   0.78 (12.00/ 15.19) &    0.78 \\
    0.25 &   1.52 &   8.90  &    7.48 (5.02/ 0.66) &     10.21 \\
    0.50 &   1.58 &   9.12  &    8.20 (3.81/ 0.46) &     12.72 \\
    0.75 &   1.64 &  37.10  &    8.98 (3.26/ 0.37) &     14.75 \\
  \bottomrule
  \end{tabular}
  \caption{Vary heuristic accuracy, fix r=0.1\%, s=512}
  \label{tab:vary-block}
\end{subtable}

\begin{subtable}{\linewidth}
  \centering
  \begin{tabular}{l|lll}
    \toprule
    & \multicolumn{3}{c}{Improvement Factors} \\
   resolution& hp-opt  & expd cost                      & runtime \\
    \midrule 
    256  &  1.31 &   2.85 (1.21/ 0.43) &      3.80 \\
    512  &  1.29 &   4.62 (3.63/ 0.79) &      5.97 \\
    1024 &  1.29 &  5.10 (11.32/ 2.24) &      6.59 \\
    2048 &  1.29 &  5.47 (38.73/ 7.12) &      7.12 \\
  \bottomrule
  \end{tabular}
  \caption{Vary resolution, scale up a map r=0.1\%, s=256, b=75\%}
  \label{tab:vary-l}
\end{subtable}

  \caption{
    Improvement factors of various metrics ($\frac{Metric(JPS)}{Metric(CJPS)}$) on three settings, $>$1 means improvements, where:
    (1) \textit{hp-opt} measures heap operations (\#expansion + \#insertion);
    (2) \textit{subopt} measures \textit{opt} on suboptimal nodes;
    (3) \textit{TPE} (time cost per expansion) measures average cost per node expansion in $\mu$s ($\frac{time}{\#expd}$),
    we also reveal the raw TPE of JPS and CJPS in parenthesis;
    (4) \textit{runtime} measures average runtime per query ($\frac{time}{\#queries}$).
  }
  \label{tab:vary-factors}
\end{table}
This experiment shows how CJPS is affected by three map properties: random-obstacles density, heuristic accuracy and resolution (size of map). 
To do this, we run queries on three sets of synthetic maps:
\begin{itemize}
  \item Fix s=512, b=75\%, vary r $\in$ \{0, 0.1, 1, 10, 20\}\%;
  \item Fix r=0.1\%, s=512, vary b $\in$ \{0, 25, 50, 75\}\%;
  \item Fix the map (r=0.1\%, s=256, b=75\%), vary resolution $\in$ \{256, 512, 1024, 2048\}.
    This affects map size but not topology; i.e., the number of jump points is the same. 
\end{itemize}
\emph{Diagonal Caching} and \emph{Backwards Scanning} \textbf{were not} applied in CJPS in this experiment.
Table~\ref{tab:vary-factors} shows the results. 
From Table~\ref{tab:vary-r}, we can see that when there is no chance to prune (r=0\%), CJPS is 21\% slower than JPS. 
This is due to the overhead of additional local reasoning in diagonal recursion.
With 0.1\% random obstacles pruning becomes effective and CJPS is 14.87 times faster than JPS.
When $r$ increases, the average expansion cost of CJPS is stable, and although there are more chances to prune (i.e., more search nodes), 
the improvement factor drops. The reason is that the g-value differences between optimal and suboptimal nodes are smaller, 
and the upper bound estimation (equation~\ref{eq:gp}) is relatively less accurate,
which weakens the pruning (i.e., \textit{subopt} drops). Meanwhile, JPS expansion cost becomes smaller as the diagonal recursion terminates earlier. 
Thus, CJPS becomes less effective and eventually slower than JPS when r=20\%.

From Table~\ref{tab:vary-block}, we can see that CJPS is more effective when the heuristic is inaccurate,
because the search space becomes larger and JPS tends to generate more suboptimal search nodes.
In Table~\ref{tab:vary-l} we see the number of heap operations (\emph{opt}) are the same on both maps. 
This is because simply scaling up doesn't change the number of jump points in the search space.
We also see the expansion cost improvement of CJPS increases, especially up to 1024. 
The reason is that scanning a higher-resolution grid map is slower, due cache behaviour.
Thus the reduced scanning in CJPS saves more time. 
As the map size grows to 2048, further improvement is diminished. 
Here CJPS has more cache misses, as the local reasoning needs to access entries in a larger 
g-value table (our table size scales with map size).

{\bf Discussion: }
CJPS can achieve significant improvements in dynamic scenarios (r$>$0\%), especially when the heuristic is inaccurate or the resolution is high.
Its improvement factor drops with the increasing density of obstacles.

\subsection{Exp-2: Ablation Study}
\begin{table}[tb]
  \small
  \begin{subtable}{\linewidth} 
  \centering
  \begin{tabular}{lrrrrrr}
  \toprule
  r(\%)&   jps &  jps-g &  jps-b &  cjps &  cjps-g &  cjps-b \\
  \midrule
  0.0 &  0.00 &   0.00 &   0.00 &  0.00 &    0.00 &    0.00 \\
  0.1 &  0.52 &   0.52 &   0.29 &  0.06 &    0.06 &    0.06 \\
  1.0 &  0.71 &   0.69 &   0.60 &  0.29 &    0.27 &    0.17 \\
  10.0&  0.59 &   0.57 &   0.50 &  0.47 &    0.43 &    0.39 \\
  20.0&  0.44 &   0.40 &   0.35 &  0.37 &    0.35 &    0.31 \\
  \bottomrule
  \end{tabular}
  \caption{Proportion of suboptimal expansion: $\frac{\sum SuboptExpd}{\sum Expd}$}
  \label{tab:variants-subopt}
\end{subtable}
\begin{subtable}{\linewidth}
  \centering
  \begin{tabular}{lrrrrr}
  \toprule
  r(\%) &  jps-g & jps-b & cjps  & cjps-g& cjps-b\\
  \midrule                                       
  0.0   &  0.97  & 0.95  & 0.75  & 0.77  & 0.77  \\
  0.1   &  0.99  & 1.02  & 14.61 & 13.55 & 11.91 \\
  1.0   &  1.00  & 0.75  & 6.32  & 6.13  & 6.14  \\
  10.0  &  0.93  & 0.85  & 1.20  & 1.18  & 1.19  \\
  20.0  &  0.93  & 0.86  & 0.98  & 0.98  & 0.96  \\
  \bottomrule
  \end{tabular}
  \caption{Speedup factor:  $\frac{avg(Time_{jps})}{avg(Time_{*})}$, where $*$ are variants: jps-g, jps-b, cjps, cjps-g, cjps-b.}
  \label{tab:variants-speedup}
\end{subtable}

  \caption{Results on synthetic maps. We look at
  suboptimal expansions and run time.}
  \label{tab:variants}
\end{table}
This experiment is to show whether it is worth further eliminate redundant work by diagonal caching (\textit{-g}) and backwards scanning (\textit{-b}).
We run both variants (\{JPS, CJPS\} $\times$ \{\textit{-g, -b}\}) on the synthetic map set that shows CJPS is less effective when the density increases (s=512, b=75\%, r $\in$ \{0, 0.1, 1, 10, 20\}\%). Table~\ref{tab:variants} shows the results.
We can see that both \textit{-g} and \textit{-b} reduce the proportion of suboptimal node expansion on top of JPS and CJPS (Table~\ref{tab:variants-subopt}),
but the improvement is not enough to pay the overhead, so they don't achieve better performance (Table~\ref{tab:variants-speedup}). 
Thus, in the next experiment, we focus on CJPS.

\subsection{Exp-3: Domain Maps}

\begin{figure*}[bth]
  \centering
  \begin{subfigure}{\linewidth}
    \centering
    \includegraphics[width=.9\linewidth]{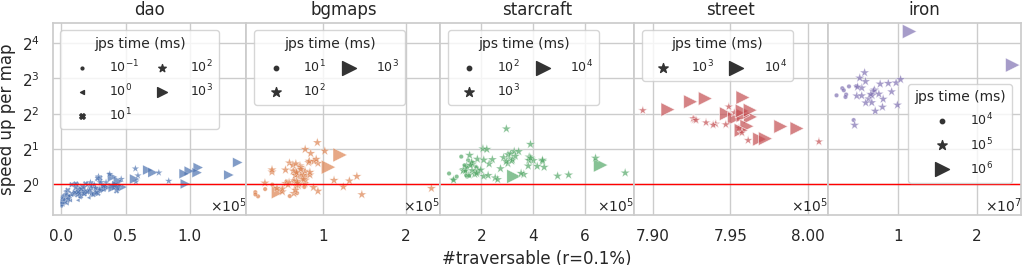}
    \caption{Speed-up per map in increasing number of traversable cells.}
    \label{fig:speed-per-map}
  \end{subfigure}
  \begin{subfigure}{\linewidth}
    \centering
    \includegraphics[width=.9\linewidth]{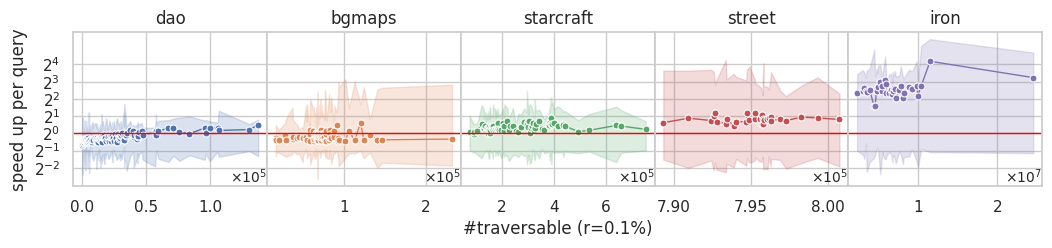}
    \caption{Compact distribution of speed-up on queries per map. Points represent the median, and the bands represent the min and max values.}
    \label{fig:dist-per-map}
  \end{subfigure}
  \begin{subfigure}{\linewidth}
    \centering
    \includegraphics[width=.9\linewidth]{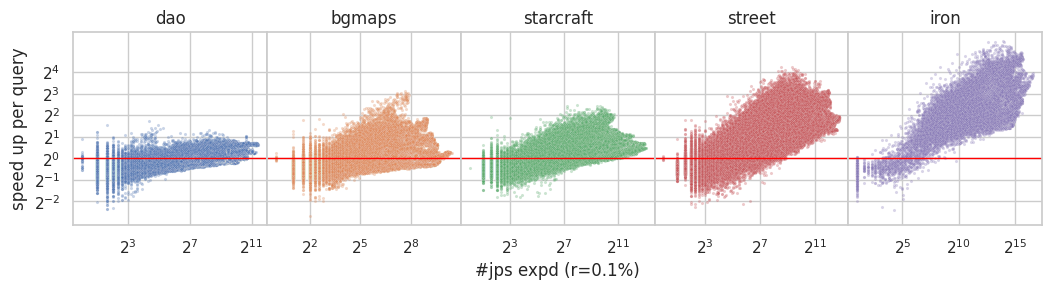}
    \caption{Speed-up per query in increasing difficulty (\textit{jps expansion})}
    \label{fig:speed-expd}
  \end{subfigure}
  \caption{Speed-up on cumulative time ($\frac{\sum Time_{jps}}{\sum Time_{cjps}}$) and queries($\frac{Time_{jps}}{Time_{cjps}}$). The red line is 1, and values above it indicate that CJPS is faster.}
  \label{fig:domain_cmp}
\end{figure*}

\begin{table}[tb]
  \centering
\begin{tabular}{llrrr}
\toprule
      & time(s)   & cjps     & jps         &  speed up \\
r (\%)& domain &         &              &           \\
\midrule
0.0   & dao &   13.93 &     13.58       &      0.97 \\
       & bgmaps &    4.95 &      3.49    &      0.71 \\
       & starcraft &   37.86 &     33.12 &      0.87 \\
       & street &   17.14 &     13.44    &      0.78 \\
      & iron &  167.35 &    175.17      &      1.05 \\
\midrule
0.1   & dao &   15.52 &     17.22 &      1.11 \\
       & bgmaps &    6.49 &      7.51    &      1.16 \\
       & starcraft &   47.20 &     68.88 &      1.46 \\
       & street &   28.09 &     98.57 &      3.51 \\
      & iron &  341.78 &   2508.75 &      7.34 \\
\bottomrule
\end{tabular}

  \caption{Cumulative time per domain on static and dynamic environments.}
  \label{tab:domain-time}
\end{table}

This experiment is to show CJPS performance on public benchmarks from real applications when environments are dynamic. 
There are three resolutions for city maps (\textit{street}). We pick the highest one (1024) as it is more challenging.

\paragraph{Simulating Dynamic Environments.} In real applications, the number of dynamic changes depends on the size of the map and is not very dense. 
For example, in Starcraft, each player controls at most 200 agents and the number of facilities is usually much smaller than this, while the maps are usually 512$\times$512.
Thus, we assume the density of random obstacles is 0.1\%. 
To simulate dynamic environments,
we add random obstacles in the same way as for the synthetic maps, assuming the result represents the map after a dynamic change.
Removing obstacles has less effect on JPS search unless we analyse the map (e.g. removing a small part of a wall has almost no effect), so we dont use it here.

Table~\ref{tab:domain-time} shows the total time to finish all queries per domain, we can see that CJPS has no advantage when the environment is static,
but wins in dynamic environments, Figure~\ref{fig:domain_cmp} shows detailed results in such environments. 

\paragraph{How to Read These Plots.} Figure~\ref{fig:speed-per-map} shows speed-up on the cumulative time per map, where markers indicate the order of time to finish all queries. 
It describes an overview of improvement but misses the distribution of improvement per query. 
Figure~\ref{fig:dist-per-map} shows a compact distribution of speed-up on queries per map, including min, median and max, but it doesn't directly show what kind of queries are improved.
Figure~\ref{fig:speed-expd} shows the query speed-up in terms of increasing difficulty, i.e., number of JPS node expansion.

We can see that in \emph{dao} and \emph{bgmaps}, there are no significant improvements. 
Even though the speed-up of cumulative time can be up to 2 on specific maps, most instances are slower than JPS. 
The reason is that most maps from these domains are either small or easy. \emph{Dao} has many small dungeon-like maps, and all maps from \emph{bgmaps} are scaled to 512 from a smaller size which have large open spaces and accurate heuristics. 
In the rest of the domains, as shown in Figure~\ref{fig:dist-per-map}, CJPS has better performance on most maps. 
In \emph{starcraft}, most queries get improved, i.e., the median point is above red line. In \emph{street}, the median speed-up factors are about 2. 
In \emph{iron}, CJPS is more than 4x faster on most maps, and can be up to 16x on one large map.
According to Figure~\ref{fig:speed-expd}, an encouraging feature is that, when it is improving CJPS becomes more effective, the harder the query is.

\section{Conclusion and Future Work}
In this paper, we study the pathological behaviours of \emph{JPS}
and propose a new approach, \emph{CJPS}, which effectively 
resolves these issues and convincingly improves JPS performance 
in dynamic environments. 
Although JPS still wins when the environment is static, 
offline-based methods should be applied in this case. 

An interesting future direction is grid-based 
pathfinding with higher dimension, e.g. 3D voxel grid map~\cite{DBLP:conf/socs/NobesHWW22} or 2D grid maps 
with temporal obstacles~\cite{DBLP:conf/aips/HuHGSS21}, in the latter, obstacles can move and environments are changing during the query.
These problems have a larger search space and more symmetries, thus there are more opportunities for
CJPS to improve over the baseline algorithm.
\bibliography{ref}
\end{document}